\documentclass[letterpaper,10pt,conference]{ieeeconf}
\IEEEoverridecommandlockouts
\usepackage{cite}
\usepackage{amsmath,amssymb}
\usepackage{booktabs}
\usepackage{capt-of}
\usepackage{multirow}
\usepackage{graphicx}
\usepackage[table]{xcolor}
\usepackage{pifont}
\usepackage{tikz}
\usetikzlibrary{arrows.meta,positioning,shapes.geometric}
\usepackage{microtype}
\usepackage{url}
\usepackage{xspace}
\usepackage{dblfloatfix}

\makeatletter
\patchcmd{\thebibliography}{\itemsep 0pt plus .5pt\relax}{\itemsep -1.0pt\relax}{}{}
\def\fnum@table{Table~\thetable}
\patchcmd{\@makecaption}{\begin{center}{\footnotesize #1}\\{\footnotesize\scshape #2}\end{center}}{\begin{center}{\footnotesize\fontfamily{ptm}\selectfont #1}\\{\footnotesize\fontfamily{ptm}\selectfont #2}\end{center}}{}{}
\makeatother

\newcommand{\se}{\mathrm{SE}(3)}

\newcommand{\cmark}{\textcolor{green!55!black}{\ding{51}}}
\newcommand{\xmark}{\textcolor{red!75!black}{\ding{55}}}

\definecolor{gold}{RGB}{255,215,0}
\definecolor{silver}{RGB}{192,192,192}
\definecolor{bronze}{RGB}{205,127,50}

\title{\LARGE \bf
G6D: Geometric Learning-Free RGB-D 6D Pose Solver for Robotic Manipulation  
}

\author{
\authorblockN{
Yixuan Liang\textsuperscript{1},
William Chen\textsuperscript{2},
Yunan Wang\textsuperscript{1},
Jizhou Yan\textsuperscript{1},
Zhao Jin\textsuperscript{1},
Changling Liu\textsuperscript{2},
and Chuxiong Hu\textsuperscript{1,*}}
\authorblockA{
\textsuperscript{1}Tsinghua University \qquad
\textsuperscript{2}Sapient Intelligence\\
\textsuperscript{*}Corresponding author: Chuxiong Hu}
}

\begin{document}
% Avoid stretching figure/text gaps merely to align column bottoms.
\raggedbottom
\maketitle
\thispagestyle{empty}
\pagestyle{empty}

\begin{abstract}
6D object pose estimation is fundamental to robotic manipulation and automation. Recent zero-shot methods have significantly improved generalization to unseen objects, but most still rely on large-scale pretrained models with substantial GPU computation and memory demands. These requirements complicate deployment on robotic platforms where perception, planning, and control share limited computational resources, while learned intermediate representations offer limited geometric interpretability for task-specific adaptation. To address these limitations, we propose G6D, a learning-free, geometry-driven RGB-D 6D pose solver. Given an RGB-D observation, an object instance mask, camera intrinsics, and a CAD model, G6D generates pose hypotheses through template-based geometric matching and refines them using silhouette and depth consistency, forming a purely geometry-driven pose estimation paradigm. This paradigm requires neither pretrained visual models nor target-specific training and preserves interpretable geometric representations throughout pose estimation. Moreover, adjustable hypothesis counts provide flexible accuracy–computation trade-offs, while a CPU-only configuration supports deployment without GPU resources. Experiments on LineMOD and five BOP19 datasets demonstrate advanced performance. Real-world pick-and-place experiments further demonstrate G6D’s applicability to robotic manipulation. The complete project is publicly available at \url{https://ai4control.github.io/G6D-Project-Page}.

\end{abstract}

\section{Introduction}
\label{sec:intro}

6D object pose estimation recovers the 3D translation and rotation of an object and is fundamental to robotic manipulation tasks. Prior studies in robotics have investigated pose estimation in cluttered scenes, robotic grasping, industrial objects, and articulated-object manipulation~\cite{MitashBB18,ZhangH22,DengXMEBF20,KleebergerH20,TianPAL20,HuangLWFJX25}.

However, the requirements of robotic systems for pose estimation extend beyond accuracy. Objects encountered in real-world manipulation are diverse and continuously changing, requiring pose estimators to generalize to novel instances without repeated data collection and retraining. Meanwhile, the inference process should preserve interpretable intermediate representations to facilitate prior incorporation, constraint construction, failure analysis, and manipulation decision-making, enabling adaptation to task-specific requirements across diverse robotic applications. Moreover, pose estimation is only one component of the perception–planning–control pipeline and must share computational resources with other modules, such as scene understanding, motion planning, and control. Substantial and relatively fixed computational and memory requirements therefore limit deployment across robotic platforms with different computational budgets. Consequently, a practical robotic pose estimator should jointly provide accuracy, generalization, interpretability, and adaptability to heterogeneous computational budgets.

\begin{figure}[t]
    \centering
    \includegraphics[width=\columnwidth]{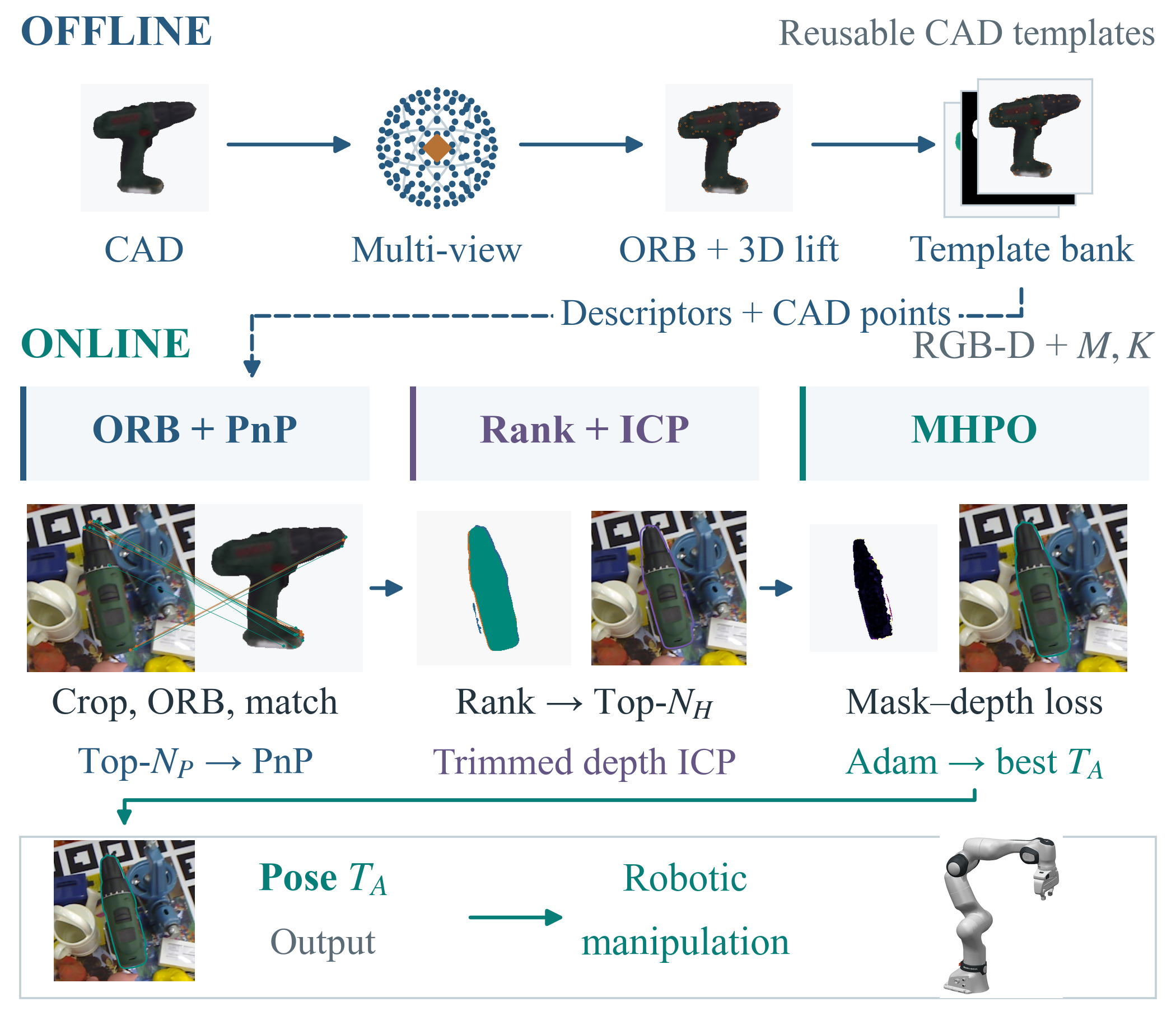}
    \caption{Overview of G6D: offline template construction and online geometry-driven pose estimation for robotic manipulation.}
    \label{fig:framework}
\end{figure}

Existing supervised RGB-D pose estimators have achieved strong performance by exploiting appearance and geometric cues. For example, DenseFusion~\cite{WangXZML0S19} fuses RGB and point-cloud features for pose prediction and iterative refinement, PVN3D~\cite{HeSHLFS20} estimates poses through 3D keypoint voting, and FFB6D~\cite{HeHFCS21} strengthens feature interaction through bidirectional RGB-D fusion. Other related approaches include PointPoseNet~\cite{Chen0BCL20}, G2L-Net~\cite{G2LNet}, and DFTr~\cite{DFTr}. However, these methods generally require object- or dataset-specific training, making additional data preparation and retraining necessary when new objects are introduced. Zero-shot RGB-D pose estimation alleviates this limitation by generalizing to unseen objects without target-specific retraining. Existing approaches typically transfer pretrained representations to novel objects and combine hypothesis generation with matching, registration, or learned refinement. FoundationPose~\cite{Wen0KB24} uses pretrained networks for pose refinement and ranking, SAM-6D~\cite{LinLLJ24} combines instance segmentation with coarse-to-fine point matching, and FreeZe~\cite{CaraffaBHP24} uses frozen geometric and visual features for 3D registration. Other related approaches include FreeZeV2~\cite{abs-2506-09784}, MatchU~\cite{Huang0YNIB24}, ZeroPose~\cite{ChenZSZWBH25}, and the depth-based OVE6D~\cite{OVE6D}.

\begin{figure*}[t]
    \vspace{2.5mm}
    \centering
    \includegraphics[width=1.0\textwidth]{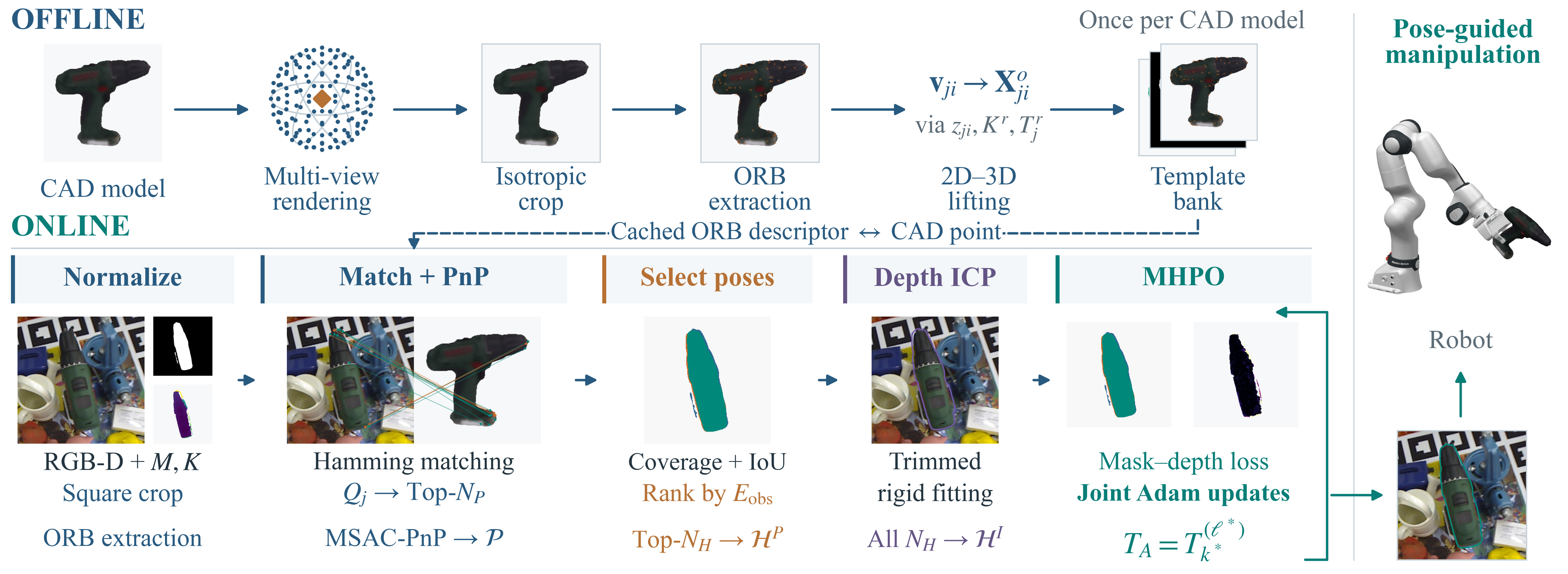}
    \caption{Detailed G6D pipeline: offline template construction, online correspondence matching and pose initialization, silhouette-guided selection, depth alignment, and mask--depth multi-hypothesis optimization.}
    \label{fig:overview}
\end{figure*}

Despite their strong accuracy and generalization to unseen objects, existing zero-shot methods still fall short of several requirements for practical robotic deployment, and zero-shot does not imply learning-free. Their reliance on large pretrained vision models limits the geometric interpretability of intermediate reasoning, while heavy and relatively fixed neural inference pipelines impose non-trivial GPU computation and memory demands. These characteristics hinder task-specific adaptation and deployment across robotic platforms with different computational budgets.

To address these limitations, this work introduces G6D, a learning-free, geometry-driven RGB-D 6D pose solver, as illustrated in Fig.~\ref{fig:framework}. The main contributions are:

\begin{enumerate}

\item \textbf{A learning-free, geometry-driven RGB-D pose estimation paradigm} that eliminates both pretrained visual models and target-specific training, while maintaining explicit geometric intermediate representations throughout inference.

\item \textbf{A geometric multi-hypothesis estimation and refinement pipeline} that addresses pose optimization under uncertain initialization by preserving multiple candidate poses and progressively refining them using silhouette and depth consistency.

\item \textbf{A flexible deployment design} that adapts to different computational budgets through adjustable hypothesis counts and a CPU-only configuration, enabling deployment across robotic platforms with heterogeneous computational resources.

\end{enumerate}

\section{Related Work}
\label{sec:related}

\subsection{RGB-D-based 6D Pose Estimation}

Existing learning-based RGB-D pose estimators exploit complementary appearance and geometric cues from RGB images and depth observations. Representative methods such as DenseFusion~\cite{WangXZML0S19}, G2L-Net~\cite{GaoLWHZF20}, and related RGB-D regression approaches~\cite{TianPAL20,KleebergerH20} directly infer object poses from fused image and point-cloud features. Other methods, including PointPoseNet~\cite{Chen0BCL20}, PVN3D~\cite{HeSHLFS20}, FFB6D~\cite{HeHFCS21}, and RCVPose~\cite{WuZEG22}, learn geometric keypoints or point-wise correspondences and recover poses through voting or rigid alignment. Although these methods achieve strong accuracy, they generally require object- or dataset-specific training, limiting direct generalization to unseen objects.

\subsection{Zero-shot RGB-D Pose Estimation}

Recent zero-shot RGB-D methods aim to estimate poses of unseen objects without target-specific training. Representative approaches include OVE6D~\cite{OVE6D}, FoundationPose~\cite{Wen0KB24}, SAM-6D~\cite{LinLLJ24}, MatchU~\cite{Huang0YNIB24}, FreeZe~\cite{CaraffaBHP24}, FreeZeV2~\cite{abs-2506-09784}, and ZeroPose~\cite{ChenZSZWBH25}. These methods typically combine CAD models or reference observations with learned visual features and geometric matching to generate and refine pose hypotheses. Despite their strong generalization and accuracy, many still rely on pretrained visual models and neural inference, leading to limited geometric interpretability and non-trivial computational demands.

\subsection{Geometry-driven Pose Estimation}

Geometry-driven pose estimation traditionally relies on hand-crafted local features such as SIFT~\cite{SIFT} and ORB~\cite{ORB} to establish 2D--3D correspondences, followed by geometric solvers such as PnP/EPnP~\cite{EPnP} with RANSAC~\cite{RANSAC} to recover object poses. More recently, differentiable rendering techniques such as nvdiffrast~\cite{nvdiffrast} have enabled image-space pose refinement~\cite{DiffDOPE}. However, existing geometry-based pipelines often achieve lower accuracy than learning-based methods and can suffer from unreliable initialization and limited robustness under challenging appearance conditions.

\section{Method}
\label{sec:method}

\subsection{Overview}
\label{sec:overview}

Given an RGB-D observation $(I,D)$ of a known target object, together with its instance mask $M$, camera intrinsics $K$, and metric CAD mesh $\mathcal{G}$, G6D estimates the object-to-camera pose $T_{c\leftarrow o}\in\se$ under a geometry-driven paradigm as shown in Fig.~\ref{fig:overview}, requiring neither task-specific training nor pretrained pose networks.

G6D follows a two-stage design with offline geometric representation construction and online multi-hypothesis pose inference. In the offline stage, multi-view RGB-D renderings are used to build an object-specific template bank containing ORB descriptors and their associated metric 3D points. During online inference, G6D matches query features against the template bank to establish 2D--3D correspondences and generates diverse pose hypotheses using MSAC-PnP. These hypotheses are then evaluated by silhouette consistency, geometrically aligned with the observed depth through trimmed ICP, and refined by mask--depth multi-hypothesis pose optimization (MHPO) using differentiable rasterization. An optional geometry-driven post-refinement stage can be activated to further improve pose accuracy when additional computation is available.

\subsection{Offline Metric Correspondence Templates}
\label{sec:template}

For each CAD object, G6D constructs a reusable template bank once offline. We sample $N_T$ templates over approximately uniform viewing directions and in-plane rotations. For template $j$, rasterization provides an RGB image $I_j^r$, mask $M_j^r$, metric depth $D_j^r$, and known pose
\begin{equation}
T_j^r=
\begin{bmatrix}
R_j^r & \mathbf{t}_j^r\\
\mathbf{0}^{\top} & 1
\end{bmatrix}
\in\se.
\end{equation}

Rendered templates and online observations share the same shape-preserving normalization. Given a mask bounding box of width $w$ and height $h$, the crop side length is $s=(1+2p)\max(w,h)$, where $p$ denotes the padding ratio. The resulting square crop is uniformly resized to $S\times S$, while the inverse affine transformation $A_j^r$ is retained to map detected features back to the original image coordinates. This normalization adjusts object scale without distorting local shape.

ORB~\cite{ORB} features are extracted from the masked and contrast-normalized template crop. Let $\mathbf{v}_{ji}$ denote the original image location of feature $i$ in template $j$, $\mathbf{d}_{ji}\in\{0,1\}^{256}$ its binary descriptor, and $z_{ji}=D_j^r(\mathbf{v}_{ji})$ its rendered depth. Since the rendering pose and metric depth are known, the corresponding CAD point is
\begin{equation}
\mathbf{X}_{ji}^{o}
=
(R_j^r)^{\top}
\left(
z_{ji}K^{-1}\widetilde{\mathbf{v}}_{ji}
-\mathbf{t}_j^r
\right).
\label{eq:template-backproject}
\end{equation}

The resulting template bank is
\begin{equation}
\mathcal{B}
=
\left\{
M_j^r,T_j^r,A_j^r,
\left\{
(\mathbf{d}_{ji},\mathbf{X}_{ji}^{o})
\right\}_{i=1}^{N_j}
\right\}_{j=1}^{N_T}.
\label{eq:template-bank}
\end{equation}

Once constructed, $\mathcal{B}$ is reused for all subsequent pose estimation of the corresponding object.
\subsection{Feature Matching and MSAC-PnP}
\label{sec:initialization}

For the query observation, G6D applies the same masking, square cropping, and contrast normalization as in the offline stage. ORB descriptors $\mathbf{d}_i^q$ are then extracted, and their feature locations are mapped back to the original image coordinates as $\mathbf{u}_i^q$.

For template $j$, each query descriptor is matched to its nearest template descriptor using Hamming distance, and matches satisfying $\delta_{ji}<\tau_H$ are retained. The resulting 2D--3D correspondence set is
\begin{equation}
\mathcal{C}_j=
\left\{
(\mathbf{u}_i^q,\mathbf{X}_{j\pi_j(i)}^o)
\mid
\delta_{ji}<\tau_H
\right\},
\label{eq:correspondences}
\end{equation}
where $\pi_j(i)$ denotes the index of the nearest template descriptor to query feature $i$, and $\delta_{ji}$ is the corresponding Hamming distance. Templates with insufficient valid correspondences are discarded.

Candidate templates are ranked using
\begin{equation}
Q_j=
\sum_{i\in\mathcal{I}_j}
\left(
1-\frac{\delta_{ji}}{\tau_H}
\right),
\label{eq:template-quality}
\end{equation}
where $\mathcal{I}_j$ denotes the retained match set. This score jointly favors correspondence quantity and descriptor quality. The top $N_P$ correspondence sets are retained for pose estimation.

For each retained set, G6D estimates a metric pose using MSAC-PnP~\cite{MSAC}. Let $\Pi_K(\cdot)$ denote perspective projection. The robust objective is
\begin{equation}
\widehat{T}_j=
\arg\min_{T\in\se}
\sum_{i\in\mathcal{I}_j}
\min
\left(
\left\|
\Pi_K(T\mathbf{X}_{j\pi_j(i)}^o)-\mathbf{u}_i^q
\right\|_2^2,
\tau_P^2
\right).
\label{eq:msac-pnp}
\end{equation}

The inlier threshold $\tau_P$ is scaled according to the projected target size and clipped to a valid pixel range, improving robustness across different image scales. Valid solutions are locally refined using the MSAC inliers, yielding the initial pose set $\mathcal{P}=\{\widehat{T}_j\}_{j=1}^{N_{\mathrm{valid}}}$.

\subsection{Observable Selection and Dense Depth Alignment}
\label{sec:observable-icp}

Low reprojection error does not necessarily imply consistency with the complete visible object. G6D therefore renders the mask $\widehat{M}(T)$ of each valid PnP hypothesis and evaluates target coverage and intersection-over-union:
\begin{equation}
C(T)
=
\frac{|\widehat{M}(T)\cap M|}{|M|},
\qquad
J(T)
=
\frac{|\widehat{M}(T)\cap M|}
{|\widehat{M}(T)\cup M|}.
\end{equation}
The observable error is defined as
\begin{equation}
E_{\mathrm{obs}}(T)
=
\omega_C[1-C(T)]
+
(1-\omega_C)[1-J(T)].
\label{eq:observable-score}
\end{equation}

Coverage suppresses hypotheses explaining only a small fraction of the target, while IoU penalizes both missing and excessive rendered regions.

The $N_H$ hypotheses with the lowest observable error form
\begin{equation}
\mathcal{H}^{P}
=
\left\{
T_k^{P}
\right\}_{k=1}^{N_H}.
\end{equation}

Metric depth is then used to refine the retained pose hypotheses. For each retained pose, the CAD model is rendered to obtain $(\widehat{M}_k,\widehat{D}_k)$. Valid observed pixels $\mathbf{u}$ inside $M$ are back-projected into
\begin{equation}
\mathcal{Y}
=
\left\{
D(\mathbf{u})K^{-1}\widetilde{\mathbf{u}}
\mid
\mathbf{u}\in M,\,
D(\mathbf{u})>0
\right\},
\end{equation}
while $\widehat{M}_k$ and $\widehat{D}_k$ define the corresponding rendered source cloud $\widehat{\mathcal{Y}}_k$.

After deterministic image-space subsampling, G6D refines each retained hypothesis using trimmed point-to-point ICP~\cite{TrimmedICP}. At each iteration, nearest-neighbor correspondences are computed, and only the closest fraction $\rho$ with residuals below
$r_{\max}=\max(r_0,\kappa_d d_{\mathcal{G}})$
is retained for rigid alignment, where $d_{\mathcal{G}}$ is the CAD diameter, $\kappa_d$ controls the object-scale-dependent tolerance, and $r_0$ provides a minimum metric cutoff for small objects. This trimming suppresses outliers caused by occlusion, mask leakage, and depth discontinuities. All $N_H$ hypotheses are refined independently, yielding the aligned set $\mathcal{H}^{I}$ for subsequent refinement.
\subsection{Mask--Depth Multi-Hypothesis Pose Optimization}
\label{sec:mhpo}

MHPO preserves the remaining pose hypotheses and refines them through differentiable rasterization~\cite{nvdiffrast}. Each hypothesis $T_k\in\mathcal{H}^{I}$ is rendered with the CAD model $\mathcal{G}$ to obtain a soft mask $\widehat{M}_k$ and metric depth $\widehat{D}_k$.

The silhouette mask loss is
\begin{equation}
\mathcal{L}_M(T_k)
=
\frac{1}{HW}
\sum_{\mathbf{u}}
\left|
\widehat{M}_k(\mathbf{u})-M(\mathbf{u})
\right|.
\label{eq:mhpo-mask}
\end{equation}

Depth consistency is evaluated only over regions where the observed target, valid measured depth, and rendered object overlap. We first define the visible-object depth scale as
\begin{equation}
z_M
=
\frac{
\sum_{\mathbf{u}}
M(\mathbf{u})
\mathbf{1}_{D>0}
D(\mathbf{u})
}{
\sum_{\mathbf{u}}
M(\mathbf{u})
\mathbf{1}_{D>0}
+\epsilon
},
\end{equation}
and the valid overlap weight as
\begin{equation}
w_k(\mathbf{u})
=
M(\mathbf{u})
\mathbf{1}_{D>0}
\widehat{M}_k(\mathbf{u}).
\end{equation}
The robust normalized depth loss is
\begin{equation}
\mathcal{L}_D(T_k)
=
\frac{
\sum_{\mathbf{u}}
w_k(\mathbf{u})
\rho_{\beta}
\left(
\frac{
\widehat{D}_k(\mathbf{u})-D(\mathbf{u})
}{
z_M
}
\right)
}{
\sum_{\mathbf{u}}
w_k(\mathbf{u})
+\epsilon
},
\label{eq:mhpo-depth}
\end{equation}
where $\rho_{\beta}$ denotes the Huber penalty. This masking prevents invalid depth measurements and non-overlapping regions from introducing spurious gradients.

The multi-hypothesis objective is
\begin{equation}
\mathcal{L}_{\mathrm{MHPO}}
=
\frac{1}{N_H}
\sum_{k=1}^{N_H}
\left[
\mathcal{L}_M(T_k)
+
\lambda_D\mathcal{L}_D(T_k)
\right].
\label{eq:mhpo-objective}
\end{equation}

Optimization and hypothesis selection are deliberately separated. Pose parameters are optimized using Eq.~\eqref{eq:mhpo-objective}, whereas hypothesis quality is evaluated using the physical consistency score
\begin{equation}
S(T_k)
=
1-J(T_k)
+
\lambda_S
\frac{
\sum_{\mathbf{u}}
w_k(\mathbf{u})
\left|
\widehat{D}_k(\mathbf{u})-D(\mathbf{u})
\right|
}{
z_M
\sum_{\mathbf{u}}
w_k(\mathbf{u})
+\epsilon
}.
\label{eq:physical-score}
\end{equation}

The depth term is a normalized weighted mean absolute residual, which prevents positive and negative depth errors from canceling. It is used only for hypothesis selection, while pose optimization uses the robust Huber loss in Eq.~\eqref{eq:mhpo-depth}.

Across all hypotheses and optimization steps, G6D retains the best state encountered:
\begin{equation}
T_A
=
\arg\min_{k,\ell}
S(T_k^{(\ell)}),
\qquad
S_A=S(T_A).
\label{eq:mhpo-output}
\end{equation}

This historical checkpoint prevents an accurate intermediate pose from being lost due to later optimization overshoot. Adam is used for pose optimization, with early termination when the best physical consistency score falls below a confidence threshold; otherwise, optimization continues to the maximum iteration budget. 

\subsection{Conditional Geometric Post-Refinement (Optional) }
\label{sec:post-refine}

The MHPO output $(T_A,S_A)$ is taken as the final estimate by default. When additional computation is available and further accuracy improvement is desired, an optional post-refinement cascade can be enabled. It first evaluates residual geometric and appearance discrepancies between the rendered CAD model and the observed RGB-D data, then explores complementary pose hypotheses from depth- and appearance-based cues, followed by the same ICP and MHPO refinement. A new estimate is accepted only when it achieves a sufficiently lower physical consistency score than $T_A$. More details can be found in the released code.

\section{Experiments}
\label{sec:experiments}

\begin{table*}[t]
\vspace{2.5mm}
\centering
\caption{RGB-D 6D pose estimation results on LineMOD using ADD(-S)@$0.1d$ (\%).
\cmark/\xmark\ indicate whether a method is free of training, target-specific
fine-tuning, and target reference RGB-D images, respectively.
Eggbox and glue ($^\ast$) use ADD-S; Avg. is the equal-object mean over all 13 objects.}
\label{tab:linemod}
\scriptsize
\setlength{\tabcolsep}{3.0pt}
\renewcommand{\arraystretch}{1.07}
\resizebox{\textwidth}{!}{%
\begin{tabular}{llccc|*{14}{c}}
\toprule
Type & Method & Train.-free & Tune-free & Ref.-free &
ape & bench & cam & can & cat & drill & duck &
egg$^\ast$ & glue$^\ast$ & hole & iron & lamp & phone & Avg.\\
\midrule

\multirow{6}{*}{\textbf{Instance}}
& DenseFusion~\cite{WangXZML0S19}
& \xmark & \xmark & \cmark
& 92.3 & 93.2 & 94.4 & 93.1 & 96.5 & 87.0 & 92.3
& 99.8 & 100.0 & 92.1 & 97.0 & 95.3 & 92.8 & 94.3\\

& PointPoseNet~\cite{Chen0BCL20}
& \xmark & \xmark & \cmark
& 97.9 & 99.6 & 98.5 & 99.4 & 99.3 & 97.5 & 96.1
& 97.9 & 100.0 & 97.8 & 99.4 & 99.1 & 98.9 & 98.4\\

& G2L-Net~\cite{G2LNet}
& \xmark & \xmark & \cmark
& 96.8 & 96.1 & 98.2 & 98.0 & 99.2 & 99.8 & 97.7
& 100.0 & 100.0 & 99.0 & 99.3 & 99.5 & 98.9 & 98.7\\

& PVN3D~\cite{HeSHLFS20}
& \xmark & \xmark & \cmark
& 97.3 & 99.7 & 99.6 & 99.5 & 99.8 & 99.3 & 98.2
& 99.8 & 100.0 & 99.9 & 99.7 & 99.8 & 99.5 & 99.4\\

& FFB6D~\cite{HeHFCS21}
& \xmark & \xmark & \cmark
& 98.4 & 100.0 & 99.9 & 99.8 & 99.9 & 100.0 & 98.4
& 100.0 & 100.0 & 99.8 & 99.9 & 99.9 & 99.7 & 99.7\\

& DFTr~\cite{DFTr}
& \xmark & \xmark & \cmark
& 98.6 & 100.0 & 100.0 & 100.0 & 100.0 & 100.0 & 99.1
& 100.0 & 100.0 & 100.0 & 99.9 & 100.0 & 99.6 & 99.8\\

\midrule

\multirow{3}{*}{\textbf{Few-shot}}
& LatentFusion~\cite{LatentFusion}
& \xmark & \cmark & \xmark
& 88.0 & 92.4 & 74.4 & 88.8 & 94.5 & 91.7 & 68.1
& 96.3 & 94.9 & 82.1 & 74.6 & 94.7 & 91.5 & 87.1\\

& FS6D~\cite{FS6D}
& \xmark & \xmark & \xmark
& 74.0 & 86.0 & 88.5 & 86.0 & 98.5 & 81.0 & 68.5
& 100.0 & 99.5 & 97.0 & 92.5 & 85.0 & 99.0 & 88.9\\

& FS6D+ICP~\cite{FS6D}
& \xmark & \xmark & \xmark
& 78.0 & 88.5 & 91.0 & 89.5 & 97.5 & 92.0 & 75.5
& 99.5 & 99.5 & 96.0 & 87.5 & 97.0 & 97.5 & 91.5\\

\midrule

& FoundationPose$^\dagger$~\cite{Wen0KB24}
& \xmark & \cmark & \cmark
& 99.0 & 100.0 & 100.0 & 100.0 & 100.0 & 100.0 & 99.4
& 100.0 & 100.0 & 99.9 & 100.0 & 100.0 & 100.0 & 99.9\\

\rowcolor{green!10}
& \textbf{G6D}
& \cmark & \cmark & \cmark
& 99.1 & 100.0 & 99.9 & 100.0 & 100.0 & 99.9 & 96.4
& 100.0 & 100.0 & 91.6 & 100.0 & 99.6 & 99.8 & 99.0\\

\rowcolor{green!10}
& \textbf{G6D (CPU-only)}
& \cmark & \cmark & \cmark
& 67.0 & 100.0 & 99.7 & 97.4 & 98.6 & 98.8 & 77.0
& 100.0 & 99.6 & 73.8 & 99.8 & 94.9 & 99.5 & 92.8\\

\rowcolor{green!10}
\multirow{-3}{*}{\textbf{Zero-shot}}
& \textbf{G6D + Post-Refine}
& \cmark & \cmark & \cmark
& 99.8 & 100.0 & 100.0 & 100.0 & 100.0 & 100.0 & 98.7
& 100.0 & 100.0 & 99.7 & 100.0 & 100.0 & 100.0
& \textbf{99.9}\\

\bottomrule
\end{tabular}}
\end{table*}

\begin{table*}[t]
\centering
\caption{Zero-shot RGB-D 6D pose estimation on five BOP19 datasets using
the official average recall (AR, \%).
AR$_5$ is the unweighted mean over the five datasets.
Mask sources are indicated in parentheses.
Test denotes dataset-provided ground-truth visible masks;
SAM-6D and NOCTIS denote predicted masks.
Training-related flags refer to the pose estimator, excluding mask generation.}
\label{tab:bop-context}
\footnotesize
\setlength{\tabcolsep}{4.5pt}
\renewcommand{\arraystretch}{1.07}

\begin{tabular*}{\textwidth}
{@{\extracolsep{\fill}}lccrrrrrr@{}}
\toprule
Method & Pretrain-free & Finetune-free &
LM-O & T-LESS & TUD-L & IC-BIN & YCB-V & AR$_5$ \\
\midrule

MatchU~\cite{Huang0YNIB24} (CNOS)
& \xmark & \cmark
& 64.4 & 52.7 & 89.8 & 44.2 & 72.6 & 64.7 \\

SAM-6D~\cite{LinLLJ24} (SAM-6D)
& \xmark & \cmark
& 69.9 & 51.5 & 90.4 & 58.8 & 84.5 & 71.0 \\

FoundationPose~\cite{Wen0KB24} (SAM-6D)
& \xmark & \cmark
& 75.6 & 64.6 & 92.3 & 50.8 & 88.9 & 74.4 \\

FreeZe~\cite{CaraffaBHP24} (SAM-6D)
& \xmark & \cmark
& 71.6 & 53.1 & 94.9 & 54.5 & 84.0 & 71.6 \\

FreeZeV2~\cite{abs-2506-09784} (SAM-6D)
& \xmark & \cmark
& 73.3 & 59.7 & 95.2 & 60.9 & 87.8 & 75.4 \\

\addlinespace[2pt]

G6D (Test)
& \cmark & \cmark
& 77.9 & 81.1 & 94.6 & 60.7 & 84.0 & 79.7 \\

G6D (SAM-6D)
& \cmark & \cmark
& 59.8 & 49.7 & 81.6 & 38.9 & 59.9 & 58.0 \\

G6D (NOCTIS)
& \cmark & \cmark
& 60.9 & 50.8 & 82.5 & 38.9 & 66.2 & 59.9 \\

\midrule
\multicolumn{9}{c}{
\textit{Additional G6D deployment/refinement variants}} \\
\midrule

G6D CPU-only (Test)
& \cmark & \cmark
& 61.0 & 73.3 & 80.2 & 49.8 & 74.7 & 67.8 \\

G6D CPU-only (SAM-6D)
& \cmark & \cmark
& 49.1 & 46.6 & 66.0 & 37.6 & 59.5 & 51.8 \\

G6D CPU-only (NOCTIS)
& \cmark & \cmark
& 49.3 & 48.2 & 68.1 & 38.7 & 64.6 & 53.8 \\

\addlinespace[2pt]

G6D + Post-refine (Test)
& \cmark & \cmark
& 80.0 & 81.4 & 94.8 & 61.9 & 83.7 & 80.3 \\

G6D + Post-refine (SAM-6D)
& \cmark & \cmark
& 61.6 & 50.5 & 83.3 & 38.6 & 59.9 & 58.8 \\

G6D + Post-refine (NOCTIS)
& \cmark & \cmark
& 63.8 & 52.7 & 85.3 & 39.2 & 66.3 & 61.5 \\

\bottomrule
\end{tabular*}

\vspace{2pt}
\parbox{\textwidth}{\footnotesize
For predicted masks, G6D uses detector-confidence top-$N$ selection;
FreeZeV2 uses $N+1$ masks with its own pose ranking.}
\end{table*}

\begin{figure*}[t]
    \centering
    \includegraphics[width=1.0\textwidth]{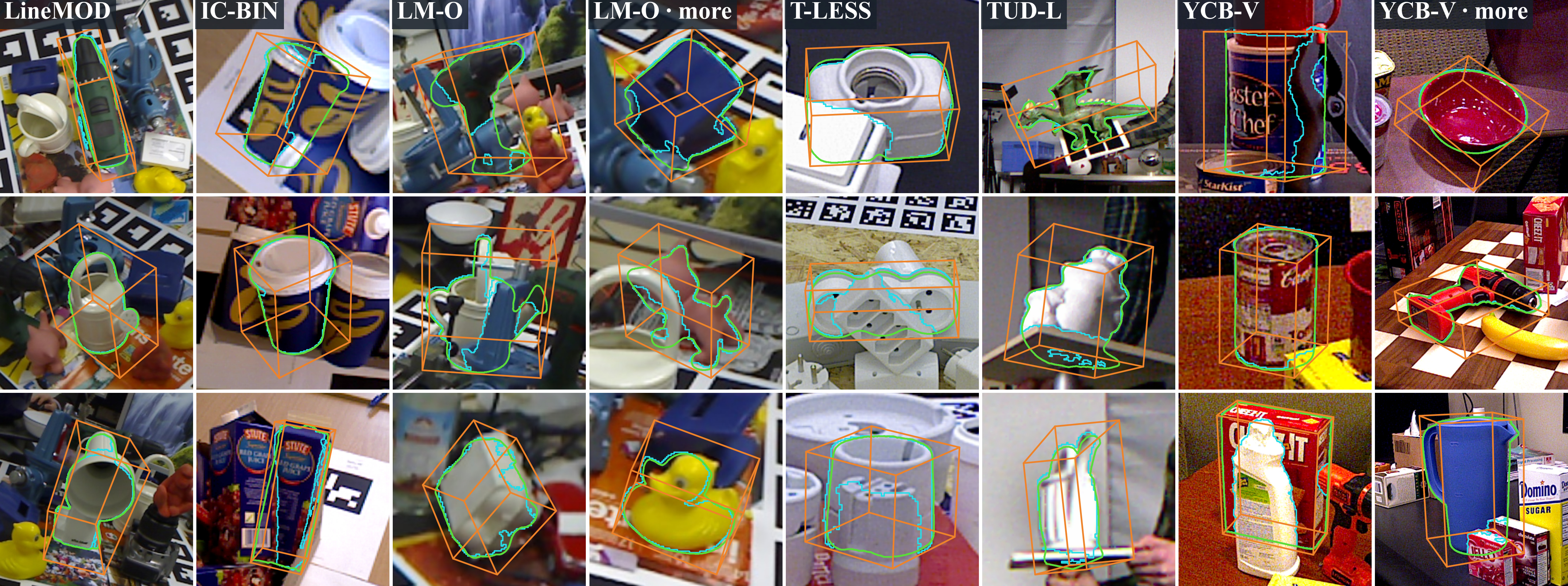}
    \caption{Qualitative 6D pose estimation results of G6D on LineMOD and five BOP19 datasets: LM-O, T-LESS, TUD-L, IC-BIN, and YCB-V. Projected 3D bounding boxes and object contours illustrate pose alignment across diverse object geometries, textures, and viewpoints, including challenging scenes with clutter and partial occlusion.}
    \label{fig:result}
\end{figure*}

\subsection{Evaluation Protocol and Metrics}
\label{sec:protocol}

We evaluate G6D on the full LineMOD test split~\cite{LineMOD} and five BOP19 datasets: LM-O, T-LESS, TUD-L, IC-BIN, and YCB-V~\cite{BOP}. For LineMOD, we report ADD for asymmetric objects and ADD-S for symmetric objects:
\begin{align}
e_{\mathrm{ADD}}
&=
\frac{1}{|\mathcal{X}|}
\sum_{\mathbf{x}\in\mathcal{X}}
\left\|
(R\mathbf{x}+\mathbf{t})-
(R^{*}\mathbf{x}+\mathbf{t}^{*})
\right\|_2,\\
e_{\mathrm{ADD\text{-}S}}
&=
\frac{1}{|\mathcal{X}|}
\sum_{\mathbf{x}\in\mathcal{X}}
\min_{\mathbf{y}\in\mathcal{X}}
\left\|
(R\mathbf{x}+\mathbf{t})-
(R^{*}\mathbf{y}+\mathbf{t}^{*})
\right\|_2 .
\end{align}
A pose is considered correct when $e_{\mathrm{ADD(-S)}}<\tau d$, where $d$ is the object diameter. We report results at $\tau=0.1$.

For BOP, we follow the official average recall metric~\cite{BOP},
\begin{equation}
\mathrm{AR}
=
\frac{1}{3}
\left(
\mathrm{AR}_{\mathrm{VSD}}
+
\mathrm{AR}_{\mathrm{MSSD}}
+
\mathrm{AR}_{\mathrm{MSPD}}
\right).
\end{equation}

G6D constructs $N_T=324$ offline templates, retains the top $N_P=256$ correspondence sets for MSAC-PnP, and preserves $N_H=18$ hypotheses for depth alignment and MHPO refinement. ORB matching uses $\tau_H=64$, observable ranking uses $\omega_C=0.70$, trimmed ICP uses $\rho=0.70$ and $\kappa_d=0.35$, and MHPO uses $\lambda_D=0.1$ and $\lambda_S=10$. Unless otherwise specified, the same configuration is used throughout all experiments.

As a pose solver, G6D directly takes an RGB-D observation, an instance mask, a CAD model, and camera intrinsics $K$ as input. Among these inputs, mask quality is particularly important because G6D is explicitly driven by geometric consistency: the mask defines the target region for feature extraction, guides silhouette-based hypothesis ranking, constrains depth alignment, and directly participates in the mask--depth objective of MHPO. Consequently, G6D can be more sensitive to mask quality than learning-based methods that implicitly encode stronger appearance priors. Although this effect is relatively limited on LineMOD, where occlusion and scene complexity are moderate, it becomes non-negligible on the more challenging BOP datasets. Therefore, evaluating G6D only with dataset-provided ground-truth masks may lead to an overly favorable comparison with modern learning-based methods that operate on predicted masks. To provide a more comprehensive and fair evaluation, we therefore report three BOP settings using: 1) the dataset-provided ground-truth masks, 2) SAM-6D~\cite{SAM6D} masks, consistent with the segmentation source used by the corresponding baseline, and 3) masks produced by the more recent NOCTIS~\cite{NOCTIS} method. 

Beyond the effect of mask quality, we further characterize G6D under different inference configurations. Specifically, we report three variants: the standard G6D pipeline, G6D with optional post-refinement, and a CPU-only variant. The CPU-only version removes GPU-dependent differentiable-rendering refinement and related stages, providing a practical reference for deployment on resource-constrained platforms without access to high-performance GPUs.

\subsection{Results and Discussion}
\label{sec:result}

Table~\ref{tab:linemod} summarizes the LineMOD results. Without pretrained models, target-specific fine-tuning, or reference RGB-D images, the standard G6D achieves 99.0\% ADD(-S) recall at $0.1d$, remaining competitive with highly optimized supervised and zero-shot approaches. The optional post-refinement further improves the average recall to 99.9\%, while the CPU-only variant still reaches 92.8\%. These results show that a purely geometry-driven pipeline can achieve high pose accuracy while retaining substantial deployment flexibility.

Table~\ref{tab:bop-context} further evaluates G6D on the more challenging BOP19 benchmarks under different mask sources. With dataset-provided masks, the standard G6D achieves 79.7\% AR$_5$, outperforming all compared zero-shot RGB-D methods while requiring neither pretraining nor target-specific fine-tuning. Post-refinement further increases AR$_5$ to 80.3\%, whereas the CPU-only variant retains 67.8\%, demonstrating that the core geometric pipeline remains effective even without GPU-dependent differentiable refinement.

With predicted masks, standard G6D obtains $58.0\%$ AR$_5$ using SAM-6D and $59.9\%$ using NOCTIS; post-refinement reaches $61.5\%$ with NOCTIS. These scores remain below the listed baselines. The difference from the supplied-mask setting motivates a distinction between missed detections, inaccurate mask boundaries, and genuine object occlusion. The results support deployment where reliable target masks are available, while limiting broader claims in cluttered scenes.

Figure~\ref{fig:result} further illustrates the estimation quality of G6D across diverse objects and scenes. Overall, the results highlight the main advantage of G6D: favorable pose accuracy can be achieved without pretrained visual representations or target-specific learning, while preserving explicit geometric intermediate representations and supporting deployment from CPU-only systems to higher-performance GPU platforms. Its current limitation lies primarily in the dependence on reliable object-region priors, making G6D most effective in scenarios where accurate masks can be readily obtained.

\subsection{Stage and Component Ablation}
\label{sec:ablation}

To ensure a consistent comparison, all results are evaluated on the same 15,800 LineMOD frames used for the main experiment.

\begin{figure}[h]
    \centering
    \includegraphics[width=0.86\columnwidth]{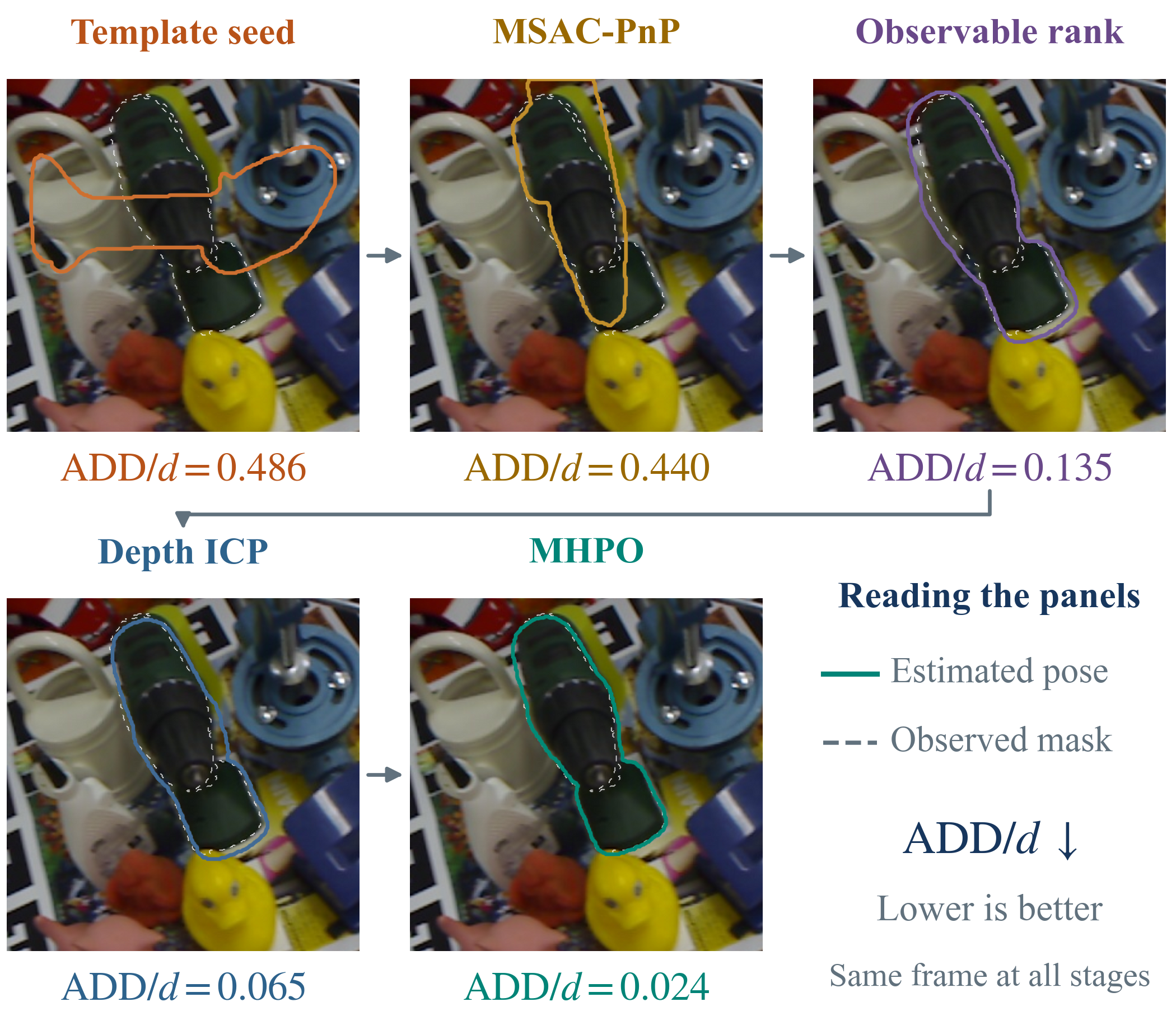}
    \caption{Stage-wise accuracy improvement along the G6D pipeline.}
    \label{fig:ablation}
\end{figure}

As illustrated in Fig.~\ref{fig:ablation} and Table~\ref{tab:stage-ablation}, G6D progressively improves pose accuracy throughout its sequential pipeline. The template seed alone achieves 8.85\% ADD(-S)@$0.1d$ recall, which increases to 33.94\% after MSAC-PnP. Observable selection provides the largest gain, improving recall by 37.86 pp to 71.80\% by filtering hypotheses according to global silhouette consistency. Subsequent depth ICP further raises recall to 90.67\%, demonstrating the effectiveness of metric depth for correcting residual geometric errors. Finally, MHPO refines the remaining hypotheses through mask--depth differentiable rendering and achieves 98.93\% recall. These results show that the overall performance arises from the progressive interaction of correspondence-based initialization, observation-consistent selection, depth alignment, and multi-hypothesis refinement.

\begin{table}[h]
\centering
\caption{Cumulative ADD(-S)@$0.1d$ recall through the five stages of the G6D
primary pipeline on all 15,800 LineMOD frames. Arrows indicate execution order.}
\label{tab:stage-ablation}
\footnotesize
\setlength{\tabcolsep}{1.5pt}
\renewcommand{\arraystretch}{1.15}

\begin{tabular*}{\columnwidth}
{@{\extracolsep{\fill}}c c c c c c c c c@{}}
\toprule

\shortstack{\textbf{Stage 1}\\Template\\seed}
& $\rightarrow$ &
\shortstack{\textbf{Stage 2}\\MSAC-PnP\\initialization}
& $\rightarrow$ &
\shortstack{\textbf{Stage 3}\\Observable\\selection}
& $\rightarrow$ &
\shortstack{\textbf{Stage 4}\\Depth\\ICP}
& $\rightarrow$ &
\shortstack{\textbf{Stage 5}\\MHPO\\refinement}
\\

\midrule

8.85\%
&&
33.94\%
&&
71.80\%
&&
90.67\%
&&
\textbf{98.93\%}
\\

\bottomrule
\end{tabular*}
\end{table}

Table~\ref{tab:component-ablation} separates the roles of selection and continuous refinement. Removing MHPO causes the largest recall drop of 8.259\%, followed by removing both ranking stages (7.253\%), observable ranking (6.127\%), and PnP initialization (3.842\%). Within MHPO, disabling pose updates, mask consistency, depth consistency, and depth-aware selection reduces recall by 1.500\%, 0.310\%, 0.608\%, and 1.101\%, respectively, while dense depth ICP contributes a smaller gain of 0.468\%.

\begin{table}[h]
\vspace{2.5mm}
\centering
\caption{Component ablation of the G6D primary pipeline on LineMOD using
ADD(-S)@$0.1d$. Drop denotes the performance decrease from the full system.}
\label{tab:component-ablation}
\footnotesize
\setlength{\tabcolsep}{4.0pt}
\renewcommand{\arraystretch}{1.08}
\begin{tabular}{lcc}
\toprule
\textbf{Configuration} & \textbf{Recall (\%)} $\uparrow$ & \textbf{Drop (pp)} $\downarrow$ \\
\midrule
\textbf{Full system}                  & \textbf{98.930} & -- \\
\midrule
w/o Template pre-ranking$^\ddagger$   & 98.816 & 0.114 \\
w/o PnP                               & 95.089 & 3.842 \\
w/o Observable ranking                & 92.804 & 6.127 \\
w/o Both ranking stages               & 91.677 & 7.253 \\
w/o Dense depth ICP                   & 98.462 & 0.468 \\
w/o Entire MHPO$^\S$                  & 90.671 & 8.259 \\
w/o MHPO pose update$^\dagger$        & 97.430 & 1.500 \\
MHPO w/o Mask loss                    & 98.620 & 0.310 \\
MHPO w/o Depth loss                   & 98.323 & 0.608 \\
Selector w/o Depth                    & 97.829 & 1.101 \\
w/o Progressive stopping              & 98.924 & 0.006 \\
\bottomrule
\end{tabular}

\vspace{2pt}
\parbox{\columnwidth}{\scriptsize
$^\dagger$A forward render is retained for hypothesis selection, but no pose
update is performed.
$^\ddagger$The same number of templates is passed to PnP, but without
quality-based pre-ranking.
$^\S$The entire MHPO block is bypassed.}
\end{table}

Template pre-ranking and progressive stopping have only marginal effects on final recall, with drops of 0.114\% and 0.006\%, respectively. Their main role is computational pruning: pre-ranking removes low-quality correspondence sets before pose solving, while progressive stopping terminates refinement once sufficient geometric consistency is reached. Both mechanisms therefore reduce unnecessary computation, making them important components of the overall pipeline.

Overall, the ablations confirm that G6D benefits from complementary geometric initialization, hypothesis selection, depth alignment, and multi-hypothesis refinement.

\subsection{Deployment Flexibility}

G6D supports flexible deployment across computational budgets by
adjusting the number of retained hypotheses $K=N_H$. The recall and peak GPU memory results in Table~\ref{tab:deployment_scalability} cover the full LineMOD test set. The reference implementation measures peak PyTorch-allocated GPU memory. Increasing $K$ consistently improves pose accuracy while increasing GPU memory usage. With only $K=2$, G6D achieves 95.39\% ADD(-S) recall with a peak GPU memory footprint of 0.13~GiB, while $K=64$ improves recall to 99.62\% at 3.38~GiB.

\begin{table}[h]
\centering
\caption{Accuracy--memory--runtime trade-off of G6D on LineMOD.}
\label{tab:deployment_scalability}
\footnotesize
\setlength{\tabcolsep}{4.0pt}
\renewcommand{\arraystretch}{1.12}

\begin{tabular}{@{}lccc@{}}
\toprule
\textbf{Budget} &
\textbf{Recall (\%)} $\uparrow$ &
\textbf{GPU Mem. (GiB)} $\downarrow$ &
\textbf{Time (s)} $\downarrow$ \\
\midrule
CPU-only & 92.78 & --   & 1.663 \\
\addlinespace[2pt]
$K=2$  & 95.39 & 0.13 & 0.914 \\
$K=4$  & 97.18 & 0.24 & \textbf{0.862} \\
$K=8$  & 98.07 & 0.45 & 0.886 \\
$K=12$ & 98.57 & 0.66 & 0.951 \\
$K=18$ & 98.93 & 0.97 & 1.067 \\
$K=24$ & 99.13 & 1.28 & 1.172 \\
$K=36$ & 99.36 & 1.91 & 1.420 \\
$K=64$ & \textbf{99.62} & 3.38 & 1.975 \\
\bottomrule
\end{tabular}
\end{table}

To assess runtime under practical deployment conditions, we use a consumer-grade personal computer with an Intel Core i9-13900HX CPU, 64~GB RAM, and an NVIDIA GeForce RTX 4090 Laptop GPU with 16~GB VRAM. Using a fixed random seed, we sample 100 distinct frames per object from four visible-mask-area quartiles, each further divided into three ordered frame-index groups. All configurations use the same 1,300 frames and supplied masks. Mean latency is weighted by stratum population within each object and then averaged equally across the 13 objects. Timing covers preprocessing and pose estimation from decoded RGB-D and mask arrays to the final pose on the CPU, with CUDA synchronization at GPU timing boundaries. Offline template construction, CUDA setup, warmup, image decoding, segmentation, and visualization are excluded.

On this personal computer, G6D achieves a mean online latency of 0.862~s at $K=4$ and 1.975~s at $K=64$. The CPU-only variant achieves 92.78\% recall without GPU resources, with a mean latency of 1.663~s. These results demonstrate deployment flexibility across CPU-only systems and GPU platforms with different memory and runtime budgets.

In addition, the explicit geometric intermediate representations allow task-specific priors, such as gravity direction, support surfaces, and restricted object orientations, to guide template and pose-hypothesis pruning, offering further opportunities to reduce computation in constrained robotic applications.

\subsection{Real-World Robotic Deployment}

\begin{figure}[htbp]
    \centering
    \includegraphics[width=\columnwidth]{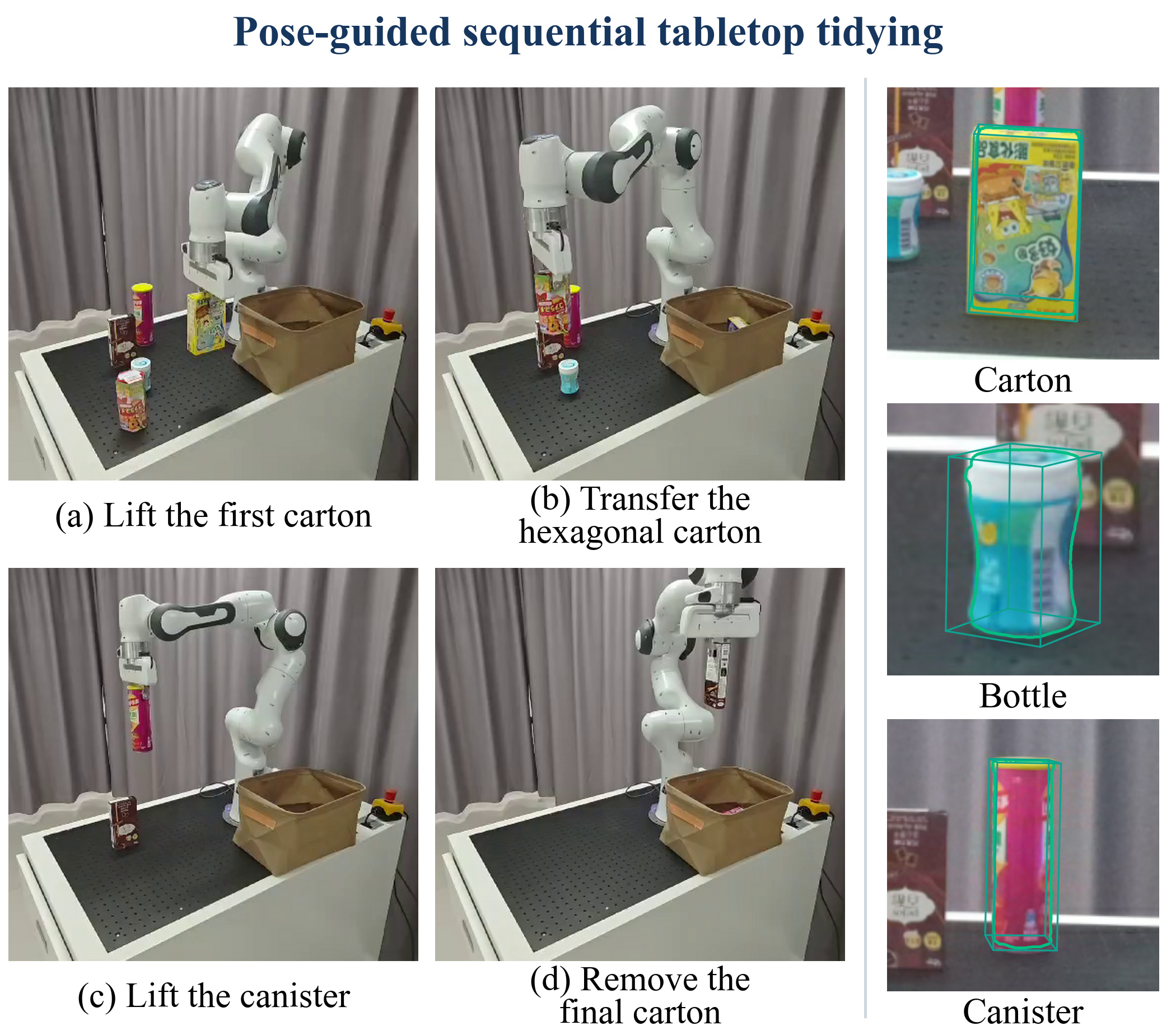}
    \caption{Pose-guided sequential tabletop tidying with G6D.
    (a)--(d) Representative stages of grasping and transferring
    objects into a storage bin.
    Bottom: estimated pose overlays for a carton, a bottle,
    and a canister.}
    \label{fig:robot_deployment}
\end{figure}

To demonstrate practical applicability, we deploy G6D in a tabletop tidying task using a Franka Emika robot and an externally mounted Azure Kinect DK RGB-D camera. The task involves grasping everyday objects with different geometric shapes and transferring them into a storage bin. LangSAM~\cite{LiuZZR23,SAM2} provides the target instance masks, which G6D uses together with RGB-D observations and CAD models to estimate object poses. In the demonstrated trials, G6D produced pose estimates suitable for guiding robotic grasping when the target masks were correctly segmented. Figure~\ref{fig:robot_deployment} shows representative manipulation stages and pose visualizations. This operator-supervised demonstration illustrates the integration
of G6D into a sequential manipulation pipeline without target-specific network training. The complete process and additional results are provided in the supplementary video.

\section{Conclusion}
\label{sec:conclusion}

This work presented G6D, a learning-free, geometry-driven RGB-D 6D pose solver. By combining geometric correspondence matching, multi-hypothesis reasoning, and mask--depth refinement, G6D provides a purely geometry-driven pose estimation paradigm and achieves favorable performance without pretrained visual models or target-specific training. Its explicit geometric intermediate representations support interpretable reasoning, while adjustable hypothesis counts and a CPU-only configuration enable deployment across diverse computational budgets. Real-world pick-and-place experiments further demonstrate its applicability to robotic manipulation, making G6D a practical solution for robotic and industrial applications.

\bibliographystyle{IEEEtranBST/IEEEtran}
\bibliography{references}

\end{document}